\documentclass{article}

\usepackage{array}
\usepackage{booktabs}
\usepackage{makecell}
\usepackage{microtype}
\usepackage{graphicx}
\usepackage{subcaption}
\usepackage{booktabs} % for professional tables

\usepackage{hyperref}

\usepackage[preprint]{icml2026}
\usepackage{natbib}

\usepackage{amsmath}
\usepackage{amssymb}
\usepackage{mathtools}
\usepackage{amsthm}

\usepackage[capitalize,noabbrev]{cleveref}

\theoremstyle{plain}

\theoremstyle{definition}

\theoremstyle{remark}

\usepackage[textsize=tiny]{todonotes}

\begin{document}

\twocolumn[
  \icmltitle{Decoupling Strategy and Execution in Task-Focused Dialogue via Goal-Oriented Preference Optimization
}



  \icmlsetsymbol{equal}{*}
  \icmlsetsymbol{corr}{†}

  \begin{icmlauthorlist}
    \icmlauthor{Jingyi Xu}{equal,nodeskai,cjlu}
    \icmlauthor{Xingyu Ren}{equal,nodeskai}
    \icmlauthor{Zhoupeng Shou}{zju}
    \icmlauthor{Yumeng Zhang}{nodeskai}
    \icmlauthor{Zhiqiang You}{corr,nodeskai}
  \end{icmlauthorlist}

  \icmlaffiliation{nodeskai}{NoDesk AI, Hangzhou, China}
  \icmlaffiliation{cjlu}{School of Information Engineering, China Jiliang University, Hangzhou, China}
  \icmlaffiliation{zju}{College of Chemical and Biological Engineering, Zhejiang University, Hangzhou, China}

  \icmlcorrespondingauthor{Zhiqiang You}{zhiqiang.you@nodeskai.com}


  \vskip 0.3in
]



\printAffiliationsAndNotice{\icmlEqualContribution. \textsuperscript{$\dagger$}Corresponding author.}
\begin{abstract}
Large language models show potential in task-oriented dialogue systems, yet existing training methods often rely on token-level likelihood or preference optimization, which poorly align with long-horizon task success. To address this, we propose \textbf{Goal-Oriented Preference Optimization (GOPO)}, a hierarchical reinforcement learning framework that decouples strategy planning from response generation via an Expert Agent and a Customer Service Agent. The Expert Agent optimizes multi-turn goal preferences at the dialogue-trajectory level, while the Customer Service Agent generates responses strictly aligned with the selected strategy.
We evaluate GOPO on public benchmarks and e-commerce customer service datasets, and introduce Task-focused Sequential Engagement (TSE), a sequence-level metric derived from real e-commerce interaction data. 
On the Mgshop dataset, GOPO improves TSE by 7.7\% and 10.3\% over PPO and Memento, with consistent gains in sequence-level reward and generation quality. 
Furthermore, a 14B model trained with GOPO achieves 2.7\% and 1.5\% higher TSE than Qwen-235B and GPT-5.2, respectively. Ablation studies confirm the Expert Agent’s critical role in long-horizon optimization. GOPO demonstrates consistent improvements across other datasets as well. This work establishes a new paradigm for task-oriented dialogue systems in commercial scenarios, with code and datasets to be made public.
\end{abstract}

\section{Introduction}
Recent advances in large language models have substantially improved task-oriented dialogue systems in fluency and language understanding~\cite{brown2020language}. However, in business-critical scenarios such as e-commerce and customer service, these systems often fall into an “efficiency trap”: despite strong technical metrics, they yield limited gains on core business KPIs, including conversion rate, user satisfaction, and first-contact resolution. This gap between linguistic proficiency and business effectiveness exposes fundamental limitations of current research paradigms~\cite{touvron2023llama}.
\begin{figure*}[htbp]
    \centering
    \includegraphics[width=0.8\textwidth]{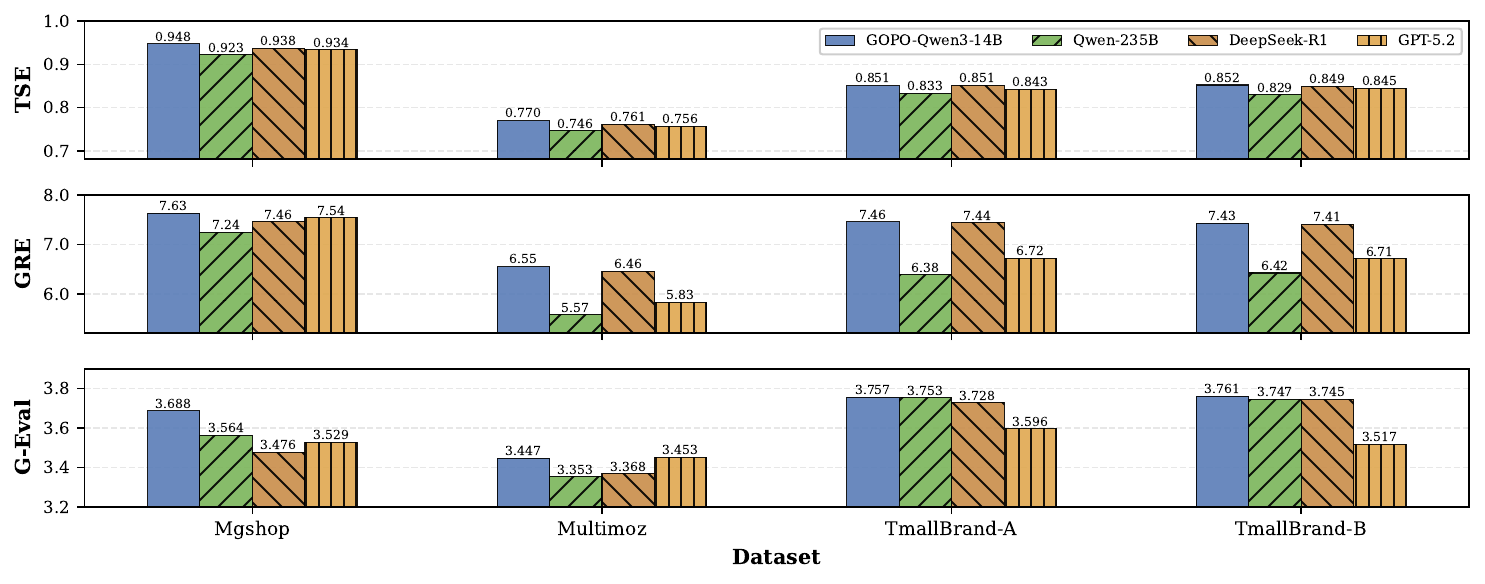}
    \caption{TSE, GRE, and G-Eval Across Datasets: GOPO-Qwen3-14B vs. Qwen-235B, DeepSeek-R1, and GPT-5.2.}
    \label{fig:my_label}
\end{figure*}
We identify three interrelated challenges underlying this issue.
First, optimization objectives are misaligned with business goals. Mainstream approaches such as supervised fine-tuning and RLHF prioritize technical criteria like preference alignment and semantic relevance, while overlooking business outcomes, resulting in fluent yet ineffective responses that fail to drive conversion or resolve complex issues~\cite{ouyang2022training}.
Second, existing single-agent architectures lack strategic flexibility in complex multi-turn dialogues~\cite{wang2025agentworkflowmemory}. By conflating strategy selection with response generation, they struggle to adapt to dynamic user intentions and emotional shifts, leading to repetitive or fragmented interactions and degraded user experience.
Finally, standard operating procedure (SOP) compliance remains unreliable. Prompt-based soft constraints cannot consistently enforce enterprise rules such as mandatory disclaimers or commitment prohibitions, causing strategic intent to be diluted during execution and leading to unstable service quality and compliance risks.

To address these challenges, we propose GOPO, a hierarchical reinforcement learning (RL) framework for task-oriented dialogue. GOPO adopts a dual-agent architecture that decouples strategy planning from response execution, comprising an Expert Agent for high-level decision-making and a Customer Service Agent for compliant response generation. The Expert Agent analyzes dialogue states and dynamically composes strategy chains from a skill pool, which are translated into structured hard constraints and enforced by the Customer Service Agent to align strategic intent with generated responses across multi-turn interactions. Moreover, GOPO employs a joint reward model grounded in business end-state metrics, aligning long-horizon optimization with real business value.
The primary contributions of this work are summarized as follows:
\begin{itemize}
  \item We propose \textbf{GOPO}, a hierarchical RL framework that decouples strategy planning from response generation via an Expert Agent and a Customer Service Agent, enabling long-horizon optimization in task-oriented dialogues.
\item We introduce \textbf{Task-focused Sequential Engagement (TSE)}, a sequence-level evaluation metric derived from real e-commerce interactions, designed to better reflect long-horizon task success.
\item Extensive experiments on public benchmarks and real-world e-commerce customer service datasets show that GOPO consistently outperforms PPO, Memento, and strong large-model baselines in sequence-level reward, generation quality, and TSE.
\item Results further demonstrate that GOPO enables smaller models to outperform significantly larger models, and ablation studies confirm the Expert Agent’s key role in long-horizon preference optimization.
\end{itemize}
The remainder of this paper is organized as follows. Section 2 reviews the related works. Section 3 presents the methodology of the GOPO framework in detail. Section 4 presents the experimental setup, results, and an in-depth analysis of the experimental findings. Finally, Section 5 concludes the paper and outlines directions for future research.

\section{Related works}
%This study is situated at the intersection of task-oriented dialogue systems, reinforcement learning–based policy optimization, and multi-agent collaboration, with a primary focus on business-driven scenarios such as e-commerce product recommendation and pre-sales services [25]. In this section, we review related work from three key dimensions: agent architectures in task-oriented dialogue, optimization methods for dialogue policies, and constraint enforcement mechanisms for SOP compliance.
\subsection{Agent Architectures in Task-Oriented Dialogue Systems}
Task-oriented dialogue systems in e-commerce serve as carriers of business value~\cite{williams2023systematic,konda2021chatbot,gamboa2023use}. Their core objective is to establish a closed loop of demand matching, trust building, and order conversion through multi-turn interactions, making this a central research focus~\cite{kanojiya2021ecommerce}. Existing approaches broadly fall into two categories: traditional rule-based or modular pipeline systems, and LLM-based end-to-end single-agent methods~\cite{martinez2024purchase}. The former rely on sequential NLU–DM–NLG pipelines, while the latter exemplified by ReAct~\cite{beyondreact2024planner} and its variants use LLMs as central controllers to implement a “think–act” loop for complex reasoning and tool use~\cite{yao2022react}.

Nevertheless, complex long-horizon business processes expose key limitations. A single agent must jointly handle SOP-driven strategy selection and compliant response generation, coupling decision logic with execution. Under uncertainty, this often results in suboptimal strategies, compliance failures, or stalled dialogues.
To overcome single-agent limitations, the community has explored multi-agent approaches. CAMEL~\cite{li2023camel} used a role-playing framework where two agents debate solutions, while other works designed agent teams with explicit role divisions, e.g., a router agent dispatching requests. Some work achieves zero-shot multi-agent coordination by training agents with a single partner across environments to learn general cooperative norms~\cite{jha2025crossenvcoop}. Although these methods partially decompose tasks, they relied on natural language soft constraints for inter-agent communication, which was unreliable in strict SOP scenarios, as LLM stochasticity can dilute strategic intent. In contrast, GOPO employs a hierarchical dual-agent architecture with structured hard-constraint transmission, ensuring Expert Agent’s strategies are executed by the Customer Service Agent.

\subsection{Optimization Methods for Dialogue Policies}
RL has long been a core approach for optimizing dialogue policies~\cite{peiyuan2024agile,wu2025aligning}. Classical methods such as Proximal Policy Optimization (PPO)~\cite{schulman2017proximal} maximized cumulative rewards, but in multi-turn dialogues rewards were sparse and delayed, with transaction success only observabled at the end of a conversation~\cite{costbench2024evaluating}. This led to severe long-horizon credit assignment issues, making optimization inefficient and prone to local optima, such as over-emphasizing fluency while neglecting SOP compliance.

To mitigate reward design challenges, preference-based optimization methods—particularly Direct Preference Optimization (DPO)~\cite{rafailov2023direct} and its variants have gained attention. DPO leveraged pairwise human preferences to bypass explicit reward modeling and online RL, and subsequent work extended this paradigm to real-time dialogue using large-scale offline data. Building on this line, Single-stream Policy Optimization (SPO)~\cite{spo2024self} and Stepwise Direct Preference Optimization (SDPO)~\cite{sdpo2024segment} introduced finer-grained supervision. However, these approaches relied on static preferences and primarily optimize technical alignment metrics, without explicitly incorporating outcome-driven business signals such as user satisfaction or purchase conversion.

Recent methods further explored long-horizon optimization by introducing trajectory-level or step-level signals. Memento~\cite{memento2024finetuning} focused on trajectory-level preference consistency, while COLLABLLM~\cite{li2025collabllm} employed multiturn-aware rewards to encourage proactive intent discovery. SWEET-RL~\cite{sweetrl2024training} addressed multi-turn credit assignment by training a critic with access to training-time information to provide step-level rewards. In contrast, GOPO adopts a joint reward model directly tied to business metrics and, through a hierarchical RL design, delivers timely supervision that effectively addresses long-horizon credit assignment.

\subsection{SOP Compliance and Constraint Enforcement}
In heavily regulated industries such as finance, healthcare, and e-commerce, strict adherence to standard operating procedures (SOPs) is critical for dialogue systems~\cite{gao2021neural}. Existing approaches mainly enforced constraints through prompt engineering or rule-based post-processing. However, prompt-based “soft constraints” are fragile to the generative flexibility of LLMs and cannot reliably ensure compliance. Post-processing filters act as passive remedies that fail to fundamentally prevent non-compliant outputs and may degrade response fluency and naturalness.

Some studies introduced stronger constraints during decoding. For instance, constrained beam search~\cite{chen2025reinforcement} and vocabulary masking~\cite{kim2025train} can enforce required keywords or block prohibited terms. While these techniques improved compliance to a limited extent, they struggled with complex logical constraints and may conflict with the model’s internal representations, resulting in degraded generation quality.

GOPO addresses these limitations through a hard-constraint transmission mechanism. The Expert Agent not only determines SOP-level strategies but also compiles them into structured hard constraints that are integrated into the Customer Service Agent’s generation process. In addition, an SOP-matching reward module provides real-time compliance feedback at the execution level, forming a closed loop from strategy planning to response generation and enabling consistent adherence to SOPs.
\begin{figure*}[htbp]
    \centering
    \includegraphics[width=1.0\textwidth]{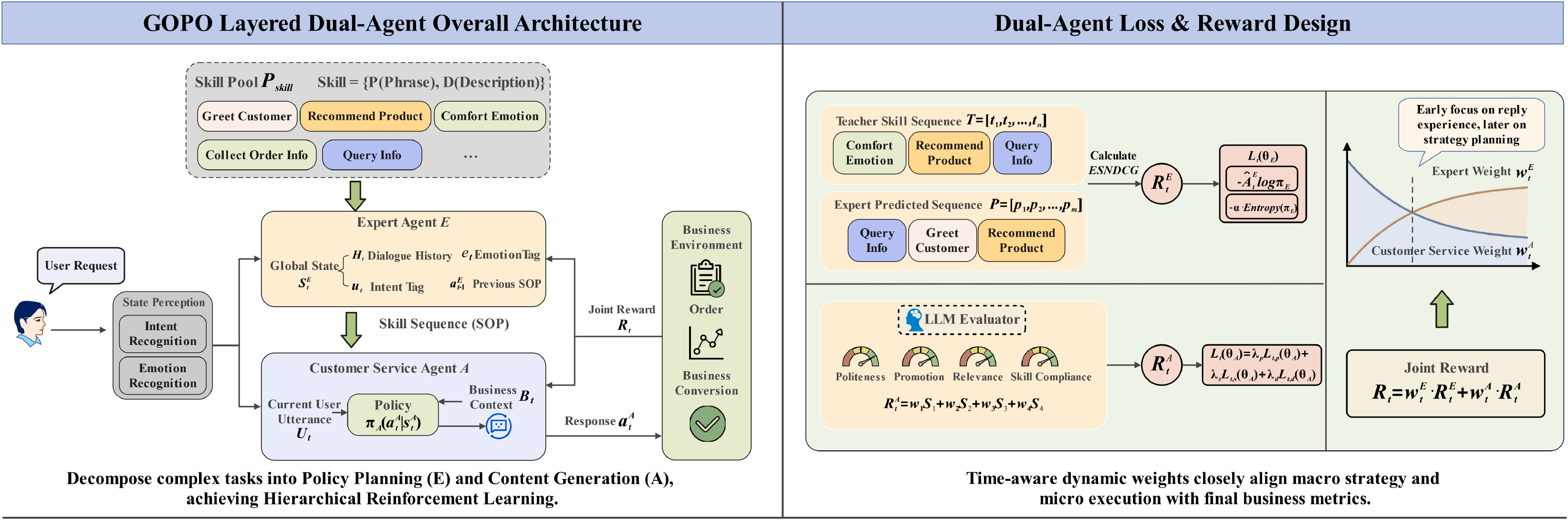}
    \caption{Hierarchical Dual-Agent Architecture of GOPO with Decoupled Strategy and Generation.}
    \label{fig:my_label}
\end{figure*}
\section{The GOPO Framework}
In this section, we introduce the GOPO framework, formalize the task-oriented dialogue problem under business constraints, and present a joint reward and loss design that aligns hierarchical RL with practical business outcomes.

\subsection{Framework Overview}
Figure~2 presents the hierarchical dual-agent architecture of GOPO.
The architecture consists of an Expert Agent for strategy planning and a Customer Service Agent for constrained response generation. Under a hierarchical RL paradigm, the Expert Agent composes skills from a predefined pool, decoupling policy planning from content generation.
\begin{enumerate}
    \item \textbf{State Perception:} The system extracts dialogue state information (e.g., user intent and emotion).
    \item \textbf{Skill Selection:} Given the global state $s_t^E$, $E$ selects a skill sequence as a macro-action $a_t^E$ to define the turn-level strategy.
    \item \textbf{Response Generation:} Conditioned on $a_t^E$ and local state $s_t^A$, $A$ generates a compliant response $a_t^A$ under hard constraints.
    \item \textbf{Feedback and Transition:} The environment returns a joint reward $R_t$ and transitions to $s_{t+1}$.
\end{enumerate}
Despite introducing hierarchical dependence, this design reduces exploration complexity and gradient variance, enabling stable long-horizon optimization.

\subsection{Problem Formalization}
We formalize the dynamic SOP task-oriented dialogue process as a Hierarchical Markov Decision Process (HMDP), represented by the tuple \[
\text{HMDP} = \langle S_E, S_A, A_E, A_A, R_E, R_A, R_{\text{total}} \rangle,
\] where independent MDPs are defined for $E$ and $A$, coordinated via hierarchical dependencies between states and actions. Each element is defined as follows:
\begin{itemize}
    \item State of Expert Agent ($s_t^E\in S_E$): Captures global and temporal dialogue information. At time $t$, the state $s_t^E$ includes:
    \begin{itemize}
        \item $H_t$: Dialogue history of the most recent $k$ turns.
        \item $u_t$: Current user intent label.
        \item $e_t$: Current user emotion label.
        \item $a_{t-1}^E$: The skill selected by $E$ to establish temporal dependency in decision-making.
    \end{itemize}
    
    \item Action ($a_t^E\in A_E$): The action space of $E$. $A^E$ is discrete, representing the selection of one or more skills from the skill pool $P_{\text{skill}}$. Therefore, $a_t^E \in P_{\text{skill}}$.
    
    \item Reward ($R_t^E\in R_E$): The reward for $E$. $R_t^E$ evaluates the long-term value of its skill choices.
    
    \item State of Customer Service Agent($s_t^A\in S_A$): $S_A$ focuses on execution details of the current turn. $s_t^A$ includes:
    \begin{itemize}
        \item $U_t$: The user’s utterance in the current turn.
        \item $a_t^E$: The skill selected by $E$, serving as a hard execution constraint.
        \item $B_t$: Relevant business context information, such as order status or inventory.
    \end{itemize}
    
    \item Action ($a_t^A\in A_A$): The action space of $A$. $A^A$ is continuous, representing the generated response text sequence. $a_t^A = (w_1, w_2, \dots, w_m)$, where $w_i$ is a token from the vocabulary $V$.
    
    \item Reward ($R_t^A\in R_A$): The reward for $A$, used to evaluate the quality and compliance of the response.
    
    \item Joint Reward ($R_t\in R_{\text{total}}$): A unified reward that jointly evaluates the strategy decision of $E$ and the response execution of $A$. This reward signal is propagated back to both agents. Detailed discussion is provided in Section 3.3.
\end{itemize}

Based on the MDP formulation, GOPO selects skills to define strategies for each dialogue turn and generates responses that satisfy immediate dialogue requirements while ensuring strict compliance and maximizing long-term business objectives.

Accordingly, GOPO adopts a hierarchical decision mechanism in which $E$ learns a macro-policy $\pi_E(a_t^E \mid s_t^E)$ to select skill combinations that guide the dialogue toward business goals, while $A$ learns a micro-policy $\pi_A(a_t^A \mid s_t^A)$ to generate high-quality, compliant responses that fit the current dialogue context. Formally, the optimization objective of GOPO is to train the joint policy $\pi = \langle \pi_E, \pi_A \rangle$ to maximize the expected cumulative reward of business outcome metrics, expressed as:
\begin{equation}
    \max_{\pi_E, \pi_A} \; \mathbb{E}_{\tau \sim \pi} \left[ \sum_{t=0}^{T} \gamma^t R_{t}(s_t^E, s_t^A,a_t^E, a_t^A) \right].
\end{equation}

\subsection{Reward and Loss Design}
The success of the GOPO framework hinges on the design of its reward and loss functions, which closely align the optimization process with actual business objectives.

The Expert Agent reward $R_t^E$ evaluates the quality of skill selection by measuring the relevance between a predicted skill sequence $P = [p_1, \dots, p_m]$ and a Teacher-generated reference sequence $T = [t_1, \dots, t_n]$. The teacher reference sequence is generated offline and serves as a stable preference anchor during training, avoiding the need for real-time oracle supervision at deployment. To quantify sequence-level relevance, we build upon Discounted Cumulative Gain (DCG)~\cite{zhang2022large}, a standard ranking-based metric that accounts for both relevance and position. For any predicted skill $s$, 
its relevance is defined as:
\begin{equation}
        \text{Rel}(s) =
\begin{cases}
n - \text{index}(s \text{ in } T) + 1, & \text{if } s \in T, \\
0, & \text{if } s \notin T.
\end{cases}
    \end{equation}
Based on this relevance definition, the DCG of the predicted skill sequence $P = [p_1, p_2, \dots, p_m]$ is computed as:
\begin{equation}
    \text{DCG}(P) = \sum_{i=1}^{m} \frac{2^{\text{Rel}(p_i)} - 1}{\log_2(i + 1)}.
\end{equation}
The Ideal DCG (IDCG) is defined as the DCG of the Teacher sequence itself, which can be expressed:
    \begin{equation}
    \text{IDCG}(T) = \sum_{j=1}^{n} \frac{2^{\text{Rel}(t_j)} - 1}{\log_2(j + 1)},
    \end{equation}
which represents the theoretical maximum score. We then define Expert Skill Normalized Discounted Cumulative Gain (ESNDCG) as a normalized variant of DCG, which is then computed as
    \begin{equation}
        \text{ESNDCG}(P, T) = \frac{\text{DCG}(P)}{\text{IDCG}(T)}.
    \end{equation}
Finally, the reward for $E$ is defined as:
\begin{equation}
R_t^E = \text{ESNDCG}(P, T).
\end{equation}

Although ESNDCG relies on teacher-generated references for reward shaping, it provides a trajectory-level preference signal rather than a supervised target, with optimization carried out via policy gradients. In practice, both the predicted and teacher skill sequence lengths are capped at 5, ensuring numerical stability while emphasizing short-horizon, high-impact skill planning.

The loss function for the Expert Agent adopts the standard policy gradient form with an added entropy regularization term to encourage exploration:
\begin{equation}
\begin{split}
\mathcal{L}_t(\theta_E) = & - \mathbb{E}_t \big[ \hat{A}_t^E \cdot \log \pi_E(a_t^E \mid s_t^E; \theta_E) \big] \\
& - \alpha \cdot \mathcal{H}(\pi_E(\cdot \mid s_t^E; \theta_E)).
\end{split}
\end{equation}

Here, $\hat{A}_t^E$ denotes the advantage function, which is computed using the Actor-Critic architecture.

The Customer Service Agent reward $R_t^A$ evaluates response quality, compliance, and diversity. We employ a large language model (GPT-4) as an automatic evaluator, which scores each response along multiple predefined dimensions. The final reward is computed as a weighted sum of these dimension scores:
\begin{equation}
    R_t^A = \alpha_1 S_1 + \alpha_2 S_2 + \alpha_3 S_3 + \alpha_4 S_4,
\end{equation}
where $\alpha_i$ are normalized weights controlling the contribution of each evaluation dimension, and $S_i$ represent the individual scores for each evaluation dimension.

The loss function for the Customer Service Agent is a composite loss, combining policy gradient, skill compliance, and diversity:

\begin{subequations}\label{eq:loss_terms}
\begin{align}
\mathcal{L}_t(\theta_A) &= 
\lambda_{\text{p}} \, \mathcal{L}_{t,\text{p}}(\theta_A) 
+ \lambda_{\text{s}} \, \mathcal{L}_{t,\text{s}}(\theta_A) + \lambda_{\text{d}} \, \mathcal{L}_{t,\text{d}}(\theta_A),\label{eq:loss_total}\\
\mathcal{L}_{t,\text{p}}(\theta_A) &= - \mathbb{E}_{a_t^A \sim \pi_A} \big[ R_t^A \, \log \pi_A(a_t^A \mid s_t^A) \big],\label{eq:loss_p}\\
\mathcal{L}_{t,\text{s}}(\theta_A) &= d(a_t^A, a_t^E),\label{eq:loss_s}\\
\mathcal{L}_{t,\text{d}}(\theta_A) &= - \sum_{t} \sum_{w \in V} p_\theta(w \mid s_t^A) \, \log p_\theta(w \mid s_t^A),\label{eq:loss_d}
\end{align}
\end{subequations}
Here, $\mathcal{L}_{t,\text{p}}$ is the policy gradient loss that maximizes the Customer Service Agent’s expected reward $R_t^A$, $\mathcal{L}_{t,\text{s}}$ enforces skill compliance via the semantic distance $d(a_t^A, a_t^E)$, and $\mathcal{L}_{t,\text{d}}$ is a negative-entropy term promoting response diversity. The total loss $\mathcal{L}_t(\theta_A)$ is their weighted sum.

To account for shifting dialogue priorities, GOPO uses a dynamic, temporally-aware weighting to combine the Expert and Customer Service Agent rewards:
\begin{equation}
    R_t = w_t^E \cdot R_t^E + w_t^A\cdot R_t^A.
\end{equation}
Here, the Expert Agent weight $w_t^E$ increases over turns, shifting emphasis from early response quality to later strategic skill selection for business goals. We adopt a temporally adaptive scheme where $w_t^E$ increases over turns while $w_t^A$ remains strictly positive, continuously enforcing response-level quality and politeness constraints.

%Through this carefully designed reward and loss structure, the GOPO framework is able to decompose abstract business objectives into concrete, learnable, and optimizable metrics for the two agents, thereby achieving end-to-end alignment from technical performance measures to business value.

\section{Experiments}
To comprehensively evaluate the effectiveness of the GOPO framework, we conducted a series of extensive experiments. This section describes the experimental setup, main results, ablations, and qualitative analysis.
\subsection{Experimental Setup}

\subsubsection{Datasets}
We evaluate our method on four datasets spanning public benchmarks and real-world enterprise data.
\begin{table*}
\centering
\caption{Comparison of GOPO (Qwen-7B-Chat) with Large-Scale Baselines Across Multiple Datasets (TSE, GRE, BLEU).}
\small
\begin{tabular}{l|*{12}{>{\centering\arraybackslash}p{0.8cm}}}
\hline
\textbf{Model} & \multicolumn{3}{c|}{\textbf{Mgshop}} & \multicolumn{3}{c|}{\textbf{Multiwoz}} & \multicolumn{3}{c|}{\textbf{TmallBrand-A}} & \multicolumn{3}{c}{\textbf{TmallBrand-B}} \\
 & TSE & GRE & BLEU & TSE & GRE & BLEU & TSE & GRE & BLEU & TSE & GRE & BLEU \\
\hline
Untrained & 0.7454 & 5.97 & 0.091 & 0.6127 & 4.53 & 0.075 & 0.6791 & 5.25 & 0.083 & 0.6685 & 5.31 & 0.082 \\
SFT & 0.8363 & 6.25 & 0.187 & 0.7015 & 4.92 & 0.149 & 0.7689 & 5.59 & 0.165 & 0.7592 & 5.64 & 0.163 \\
PPO & 0.8584 & 7.09 & 0.190 & 0.7496 & 5.68 & 0.169 & 0.8040 & 6.39 & 0.181 & 0.8126 & 6.43 & 0.183 \\
Memento & 0.8381 & 7.13 & 0.188 & 0.7205 & 5.75 & 0.153 & 0.7793 & 6.44 & 0.168 & 0.7848 & 6.38 & 0.169 \\
\textbf{GOPO(Ours)} & \textbf{0.9243} & \textbf{7.38} & \textbf{0.211} & \textbf{0.7543} & \textbf{6.38} & \textbf{0.172} & \textbf{0.8393} & \textbf{6.88} & \textbf{0.192} & 
\textbf{0.8417} & \textbf{6.92} & \textbf{0.193} \\ 
\hline
\end{tabular}
\label{tab:baseline_comparison}
\end{table*}
\begin{table*}
\centering
\caption{Comparison of GOPO with Large-Scale Baselines Across Multiple Datasets (TSE, GRE, BLEU).}
\small
\begin{tabular}{l|*{12}{>{\centering\arraybackslash}p{0.65cm}}} % 后12列宽度改为1cm
\hline
\textbf{Model} & \multicolumn{3}{c|}{\textbf{Mgshop}} & \multicolumn{3}{c|}{\textbf{Multiwoz}} & \multicolumn{3}{c|}{\textbf{TmallBrand-A}} & \multicolumn{3}{c}{\textbf{TmallBrand-B}} \\
 & TSE & GRE & BLEU & TSE & GRE & BLEU & TSE & GRE & BLEU & TSE & GRE & BLEU \\
\hline
Qwen-235B & 0.9227 & 7.24 & 0.182 & 0.7462 & 5.57 & 0.165 & 0.8327 & 6.38 & 0.114 & 0.8294 & 6.42 & 0.166 \\
DeepSeek-R1 & 0.9381 & 7.46 & 0.143 & 0.7609 & 6.46 & 0.138 & \textbf{0.8512} & 7.44 & 0.141 & 0.8489 & 7.41 & 0.139 \\
GLM-4.7 & 0.9365 & 7.25 & 0.150 & 0.7598 & 6.13 & 0.148 & 0.8487 & 7.09 & 0.153 & 0.8301 & 6.36 & 0.151 \\
GPT-5.2 & 0.9338 & 7.54 & 0.097 & 0.7561 & 5.83 & 0.092 & 0.8426 & 6.72 & 0.095 & 0.8453 & 6.71 & 0.094 \\
Gemini-2.5 & 0.9287 & 7.35 & 0.133 & 0.7451 & 6.16 & 0.129 & 0.8084 & 6.91 & 0.131 & 0.8116 & 6.88 & 0.128 \\
\hline
\textbf{GOPO-Qwen3-14B(Ours)} & \textbf{0.9475} & \textbf{7.63} & \textbf{0.279} & \textbf{0.7699} & \textbf{6.55} & \textbf{0.177} & 0.8511 & \textbf{7.46} & \textbf{0.186} & \textbf{0.8524} & \textbf{7.43} & \textbf{0.198} \\
\hline
\end{tabular}
\label{tab:large_model_comparison}
\end{table*}
\begin{table*}
\centering
\caption{Comparison of Flagship Models Across Datasets on G-Eval Scores.}
\small
\setlength{\tabcolsep}{2pt} % 默认是 6pt，减小列间距
\begin{tabular}{lcccc}
\hline
\textbf{Model} & \textbf{Mgshop} & \textbf{Multiwoz} & \textbf{TmallBrand-A} & \textbf{TmallBrand-B} \\
\hline
Qwen-235B     & 3.564 & 3.353 & 3.753 & 3.747 \\
DeepSeek-R1   & 3.476 & 3.368 & 3.728 & 3.745 \\
GLM-4.7       & 3.672 & 3.381 & 3.525 & 3.510 \\
GPT-5.2       & 3.529 & 3.453 & 3.596 & 3.517 \\
Gemini-2.5    & 3.680 & 3.407 & 3.653 & 3.534 \\
\hline
\textbf{GOPO-Qwen3-14B(Ours)} & \textbf{3.688} & \textbf{3.447} & \textbf{3.757} & \textbf{3.761} \\
\hline
\end{tabular}
\label{tab:g-eval_comparison}
\end{table*}
\begin{table*}[t]
\centering
\caption{Comparison of GOPO Variants on Multiple Datasets (TSE, GRE, BLEU).}
\small
\begin{tabular}{l|*{12}{>{\centering\arraybackslash}p{0.7cm}}}
\hline
\textbf{Model} & \multicolumn{3}{c|}{\textbf{Mgshop}} & \multicolumn{3}{c|}{\textbf{Multiwoz}} & \multicolumn{3}{c|}{\textbf{TmallBrand-A}} & \multicolumn{3}{c}{\textbf{TmallBrand-B}} \\
 & TSE & GRE & BLEU & TSE & GRE & BLEU & TSE & GRE & BLEU & TSE & GRE & BLEU \\
\hline
GOPO-Qwen-7B-Chat & 0.9243 & 7.38 & 0.211 & 0.7543 & 6.38 & 0.172 & 0.8393 & 6.88 & 0.192 & 0.8417 & 6.92 & 0.193 \\
GOPO-Qwen3-8B      & 0.9375 & 7.44 & 0.242 & 0.7680 & 6.51 & 0.175 & 0.8507 & 7.21 & \textbf{0.202} & 0.8515 & \textbf{7.43} & \textbf{0.208} \\
\textbf{GOPO-Qwen3-14B}     & \textbf{0.9475} & \textbf{7.63} & \textbf{0.279} & \textbf{0.7699} & \textbf{6.55} & \textbf{0.177} & \textbf{0.8511} & \textbf{7.46} & 0.186 & \textbf{0.8524} & \textbf{7.43} & 0.198 \\
\hline
\end{tabular}
\label{tab:GOPO_metrics}
\end{table*}
\begin{figure*}[t]
    \centering
    \begin{subfigure}{0.28\linewidth}
        \centering
        \includegraphics[width=\linewidth]{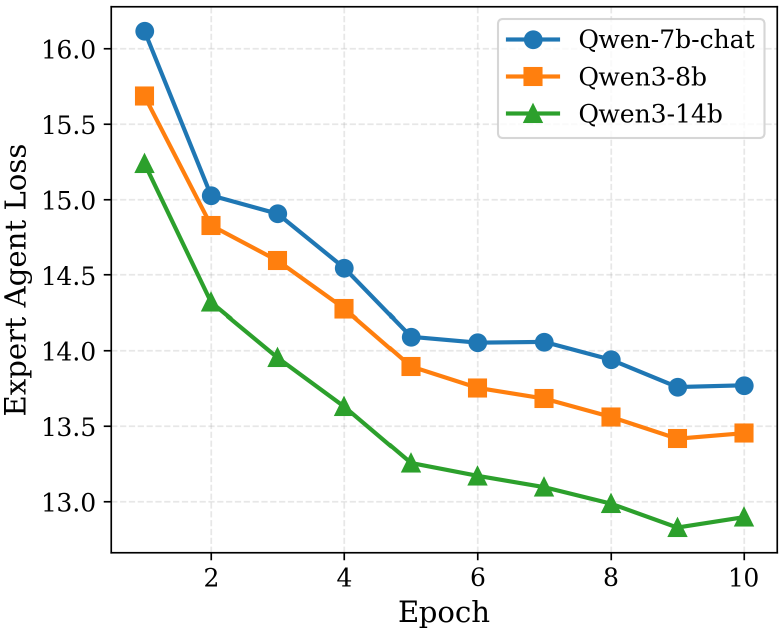}
        \label{fig:sub1}
    \end{subfigure}
    \hspace{-0.3em}
    \begin{subfigure}{0.28\linewidth}
        \centering
        \includegraphics[width=\linewidth]{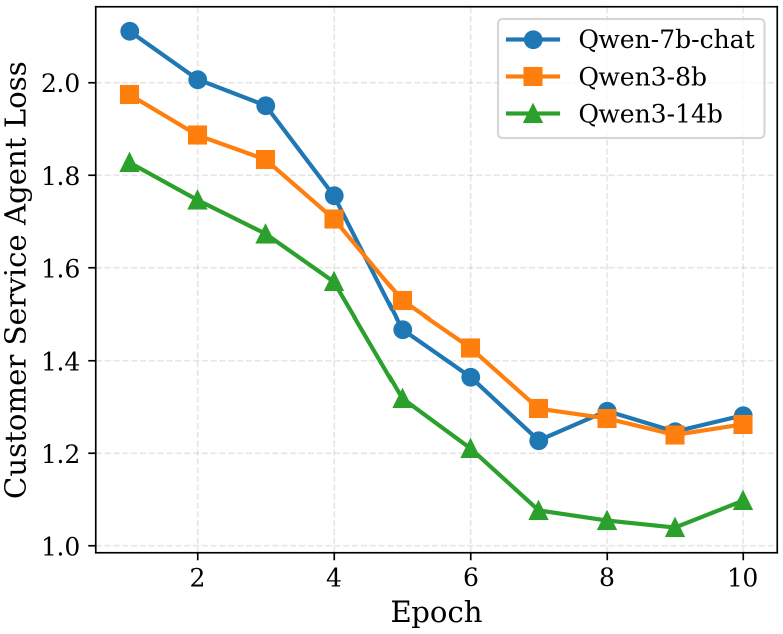}
        \label{fig:sub2}
    \end{subfigure}
    \hspace{-0.3em}
    \begin{subfigure}{0.28\linewidth}
        \centering
        \includegraphics[width=\linewidth]{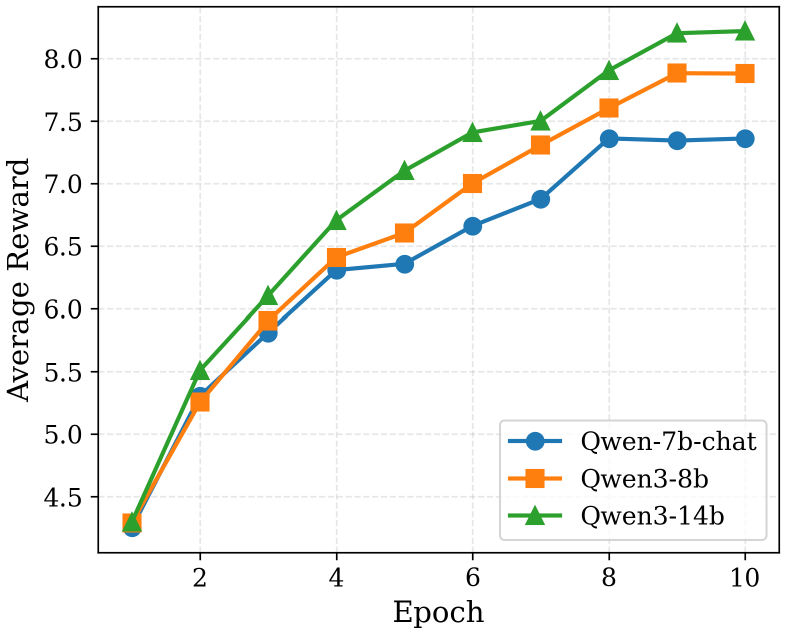}
        \label{fig:sub3}
    \end{subfigure}
    \caption{Convergence Performance of GOPO Under Different Parameter Settings.}
    \label{fig:three_subfigures_horizontal}
\end{figure*}
\begin{itemize}
    \item Mgshop~\cite{JannachDietmar2021A}: An internal test set designed to evaluate robustness under highly dynamic user emotions and rapidly changing intents.
    \item Multiwoz~\cite{budzianowski2018multiwoz}: A widely used multi-domain task-oriented dialogue benchmark for assessing general dialogue capabilities.
    \item TmallBrand-A: An internal e-commerce customer service dataset reflecting complex interactions such as purchasing, inquiries, and refunds.
    \item TmallBrand-B: An internal e-commerce customer service dataset from a different store, used to assess cross-domain generalization.
\end{itemize}
\subsubsection{Evaluation Metrics}
We adopt a multi-dimensional evaluation framework to assess performance in sequential, conditional recommendation tasks, including:
\begin{itemize}
    \item \textbf{Task-focused Sequential Engagement Success (TSE):} 
    A sequence-level metric measuring task completion efficiency in task-focused dialogues, covering requirement matching, information delivery, and proactive guidance. 
    The TSE score is defined as:
    \begin{equation}
    \begin{split}
    \text{TSE}(k) = & \ \omega_1 I_1 \gamma^{N_1 - 1} 
    + \omega_2 I_2 \gamma^{N_2 - N_1 - 1} \\
    & + \omega_3 I_3 \gamma^{N_3 - N_2 - 1}.
    \end{split}
    \end{equation}
    Here, $\omega_i$ are task weights, $I_i$ indicate task completion, $\gamma$ is a turn-based decay factor, and $N_i$ denote task completion turns. TSE measures whether key tasks are completed efficiently.

    \item \textbf{Goal-oriented Response Effectiveness (GRE):} 
    A response-level metric evaluating how individual replies advance business objectives, computed by averaging LLM-based turn-level scores on politeness, product appropriateness, and action guidance.

    \item \textbf{BLEU~\cite{papineni2002bleu}:} 
    Measures n-gram overlap with reference responses, capturing surface-level lexical similarity.

    \item \textbf{G-Eval~\cite{liu2023geval}:} 
    An LLM-based metric assessing dialogue fluency and user experience via coherence, completeness, and emotional appropriateness.
\end{itemize}

Overall, these metrics capture complementary aspects of performance across multiple levels, ranging from sequence-level goal achievement to response-level goal advancement, user experience, and surface-level similarity.

\subsubsection{Baseline Models and Comparative Experiments}
We conducted two groups of comparative experiments:
\begin{itemize}
    \item \textbf{Experiment 1: Baseline Comparison.} We compare GOPO with untrained LLMs, SFT, PPO, and Memento as representative supervised and RL baselines.

    \item \textbf{Experiment 2: Comparison with Large Models.} We benchmark GOPO against proprietary and open-source flagship models, including GPT-5.2 and Gemini-2.5 (closed-source), as well as Qwen-235B, DeepSeek-R1, and GLM-4.7 (open-source). We also evaluate GOPO implementations on Qwen-7B-Chat, Qwen3-8B, and Qwen3-14B to assess scalability across model sizes.
\end{itemize}
% In our GOPO framework, the Expert Agent and the Customer Service Agent share the same model backbone but have independent policy heads at the top layer. We use the AdamW optimizer with a learning rate of $1 \times 10^{-5}$. All experiments are conducted on servers equipped with high-performance GPUs.

\subsection{Main Results and Results Analysis}

\subsubsection{Advantages of GOPO over Mainstream Training Frameworks}

\begin{table*}[t]
\centering
\caption{Ablation Study of GOPO-14B on Mgshop and TmallBrand-A Datasets.}
\small
\begin{tabular}{l|ccc|ccc}
\hline
\textbf{Model} & \multicolumn{3}{c|}{\textbf{Mgshop}} & \multicolumn{3}{c}{\textbf{TmallBrand-A}} \\
 & TSE & GRE & BLEU & TSE & GRE & BLEU \\
\hline
Full GOPO & \textbf{0.9475} & \textbf{7.63} & \textbf{0.279} & \textbf{0.8511} & \textbf{7.46} & 0.186 \\
GOPO w/o Expert Agent & 0.9371 & 7.16 & 0.217 & 0.8023 & 6.47 & \textbf{0.227} \\
Full Baseline & 0.7353 & 6.92 & 0.251 & 0.6371 & 6.43 & 0.213 \\
Baseline w/o Expert Agent & 0.4327 & 6.08 & 0.240 & 0.1818 & 5.19 & 0.182 \\
\hline
\end{tabular}
\label{tab:ablation_14b}
\end{table*}
\begin{table*}[!htbp]
\centering
\small
\caption{Performance of Customer Service Agents Evaluated with Trained GOPO Expert Agent on Mgshop and TmallBrand-A Datasets.}
\begin{tabular}{l|*{6}{>{\centering\arraybackslash}p{0.9cm}}}
\hline
\textbf{Customer Service Agent} & \multicolumn{3}{c|}{\textbf{Mgshop Dataset}} & \multicolumn{3}{c}{\textbf{TmallBrand-A Dataset}} \\
\cline{2-7}
 & TSE & GRE & BLEU & TSE & GRE & BLEU \\
\hline
GOPO-Qwen3-14B & \textbf{0.9475} & \textbf{7.63} & \textbf{0.279} & \textbf{0.8511} & \textbf{7.46} & 0.186 \\
Qwen-235B      & 0.9352 & 7.45 & 0.214 & 0.8428 & 7.37 & 0.201 \\
DeepSeek-R1    & 0.9453 & 7.55 & 0.242 & 0.8503 & 7.03 & \textbf{0.245} \\
Gemini-2.5     & 0.9268 & 7.52 & 0.220 & 0.8025 & 6.83 & 0.131 \\
\hline
\end{tabular}
\label{tab:gopo_expert_multi_datasets}
\end{table*}

\begin{table*}[!htbp]
\centering
\small
\caption{Performance of Fixed GOPO Customer Service Agent with Different Expert Agents on Mgshop and TmallBrand-A Datasets.}
\begin{tabular}{l|*{6}{>{\centering\arraybackslash}p{1.0cm}}}
\hline
\textbf{Expert Agent} & \multicolumn{3}{c|}{\textbf{Mgshop Dataset}} & \multicolumn{3}{c}{\textbf{TmallBrand-A Dataset}} \\
\cline{2-7}
 & TSE & GRE & BLEU & TSE & GRE & BLEU \\
\hline
GOPO-Qwen3-14B & \textbf{0.9475} & \textbf{7.63} & \textbf{0.279} & \textbf{0.8511} & \textbf{7.46} & 0.186 \\
Qwen-235B      & 0.8714 & 7.40 & 0.193 & 0.8291 & 6.27 & 0.203 \\
DeepSeek-R1    & 0.7854 & 6.73 & 0.234 & 0.8395 & 6.84 & 0.146 \\
Gemini-2.5     & 0.7384 & 5.30 & 0.251 & 0.8163 & 6.55 & \textbf{0.228} \\
\hline
\end{tabular}
\label{tab:fixed_customer_multi_experts}
\end{table*}
Across all four datasets, GOPO consistently outperforms the untrained model and strong baselines including SFT, PPO, and Memento. Specifically, GOPO improves TSE by \textbf{23\%--26\%} and GRE by \textbf{23\%--40\%} over the untrained model, demonstrating the effectiveness of goal-oriented training.

Compared to SFT, GOPO achieves \textbf{7.5\%--10\%} higher TSE, highlighting its advantage in sequence-conditioned, business-driven dialogue tasks beyond token-level NLG optimization. Relative to PPO, GOPO yields \textbf{0.6\%--7.7\%} gains in TSE and \textbf{3\%--20\%} in GRE, validating the benefit of dual-agent decomposition and business-aligned GRE design over single-agent RL. Finally, GOPO outperforms Memento by \textbf{9\%--12\%} in TSE and \textbf{3\%--8\%} in GRE, indicating strong robustness across training paradigms.
\subsubsection{Comparison with Large-Scale Frontier Models}
We compare GOPO with large-scale frontier models, including Qwen-235B, DeepSeek-R1, GLM-4.7, GPT-5.2, and Gemini-2.5, and observe three key findings.

First, GOPO achieves competitive performance with significantly fewer parameters. Despite being over an order of magnitude smaller (7B/8B vs.\ 235B+), GOPO-Qwen-7B and GOPO-Qwen3-8B attain comparable TSE scores. On Mgshop, GOPO-Qwen3-8B reaches a TSE of \textbf{0.9375}, closely matching DeepSeek-R1 (\textbf{0.9387}).

Second, SOP-based prompt constraints yield inconsistent gains for large models. While marginal improvements are observed for some models (e.g., Qwen-235B and GLM-4.7), others (e.g., Gemini-2.5 on Multiwoz) exhibit degraded performance, indicating the brittleness and model-dependence of prompt-level soft constraints.

Finally, GOPO demonstrates superior stability across datasets and settings. Its hierarchical dual-agent architecture and explicit constraint propagation ensure reliable enforcement of business objectives, outperforming prompt-based approaches in both robustness and consistency.

Overall, these results demonstrate that GOPO provides a more effective optimization paradigm for complex, Goal-Oriented dialogue tasks under realistic business constraints.

\subsubsection{Performance Analysis under Different GOPO Parameter Settings}
The training trends and experimental results of the GOPO framework under different parameter settings are shown in Figure~3 and Table~4. It can be observed that as the model scale increases, GOPO achieves improvements in TSE, GRE, and BLEU across all datasets, with GOPO-Qwen3-14B performing the best. The medium-sized GOPO-Qwen3-8B also significantly outperforms GOPO-Qwen-7B-Chat, while the performance gains vary across datasets, with more noticeable improvements on Mgshop and the TmallBrand series. It is worth noting that the loss value of the Expert Agent remains relatively high and decreases only marginally during training. This behavior is expected given that the Expert Agent is optimized with a policy-gradient objective augmented by entropy regularization, where the loss magnitude reflects both non-trivial advantage estimates and sustained policy stochasticity rather than convergence to zero. Notably, despite the Expert loss stabilizing at a higher value, the overall reward converges reliably across all settings, indicating effective long-horizon optimization.
\subsection{Ablation Study}
To assess the contribution of each component in GOPO, we conduct ablation studies on Mgshop and TmallBrand-A. As shown in Table 5, we compare the full GOPO framework with a variant without the Expert Agent, as well as the corresponding untrained baseline.

Removing the Expert Agent causes substantial performance degradation, confirming its critical role in long-horizon optimization. On TmallBrand-A, TSE drops by \textbf{5.7\%} and GRE decreases by \textbf{13.3\%}. Notably, although BLEU slightly increases, TSE declines sharply, indicating that without strategic guidance the model produces conservative yet inefficient responses. This further exposes the limitation of BLEU as a surface-level metric for task-oriented dialogue.

Compared to the untrained baseline, Full GOPO achieves significant gains, improving TSE by \textbf{28.9\%} and \textbf{33.6\%}, and GRE by \textbf{10.3\%} and \textbf{16.0\%} on the two datasets. These results demonstrate that the ESNDCG-driven hierarchical RL objective effectively guides the model toward efficient task completion.

Interestingly, even without specialized training, the dual-agent architecture alone yields clear improvements over the single-agent baseline, suggesting that decoupling strategy planning from response generation better matches the structure of task-oriented dialogue.

Finally, mixed-agent experiments in Tables 6 and 7 show consistent performance drops when either agent is replaced by a non-jointly trained counterpart, confirming that GOPO’s effectiveness arises from the synergistic optimization of both agents rather than any single component.

\section{Conclusion}
This paper addresses misalignment in task-oriented dialogue systems under complex business scenarios. We propose GOPO, a hierarchical RL framework with dual agents, separating strategy planning from response generation. The Expert Agent chooses sequences of skills, while the Customer Service Agent executes responses following strict compliance, guiding the system toward desired business outcomes.
GOPO combines SOP compliance with a joint reward over entire dialogues, shifting optimization from preference imitation to direct business value maximization. Experiments on multiple real-world datasets show GOPO consistently outperforms strong baselines, including large proprietary models on TSE and GRE metrics. Current evaluation partly relies on offline TSE from real dialogues, with ongoing online tests confirming effectiveness and stability. 
Future work includes automated policy and skill discovery, expanding the skill pool dynamically, and adapting across domains. Overall, GOPO provides a practical and extensible framework for task-oriented dialogue systems and opens new directions for optimizing large multi-agent models toward specific objectives.

\newpage
% \clearpage
\bibliography{example_paper}
\bibliographystyle{icml2026}

\end{document}